\pdfoutput=1

\documentclass[11pt]{article}

\usepackage{acl}

\usepackage{times}
\usepackage{latexsym}
\usepackage{graphicx}
\usepackage{xspace}
\usepackage{amsmath}
\usepackage[noabbrev]{cleveref}
\usepackage[T1]{fontenc}

\usepackage[utf8]{inputenc}

\usepackage{microtype}

\usepackage{inconsolata}

\newcommand{\hotpot}{HotpotQA\xspace}
\newcommand{\musique}{MuSiQue\xspace}
\newcommand{\twowiki}{2WikiQA\xspace}
\newcommand{\fever}{FEVER\xspace}
\newcommand{\llama}{LLaMA\xspace}
\newcommand{\llamasize}[1]{LLaMA #1B\xspace}
\newcommand{\roberta}{RoBERTa\xspace}
\newcommand{\dragon}{DRAGON\xspace}
\newcommand{\thiswork}{Our Approach\xspace}

\newenvironment{itemizesquish}{\begin{list}{\labelitemi}{\setlength{\itemsep}{-0.2em}\setlength{\labelwidth}{0.5em}\setlength{\leftmargin}{\labelwidth}
\addtolength{\leftmargin}{\labelsep}}}{\end{list}}

\newenvironment{enumeratesquish}{\begin{list}{\addtocounter{enumi}{1}\labelenumi}{\setlength{\itemsep}{0em}\setlength{\labelwidth}{0.5em}\setlength{\leftmargin}{\labelwidth}\addtolength{\leftmargin}{\labelsep}}}{\end{list}\setcounter{enumi}{0}}

\title{Few-Shot Data Synthesis for Open Domain Multi-Hop Question Answering}

\author{Mingda Chen \qquad Xilun Chen \qquad Wen-tau Yih \\
        Meta\\
        \texttt{\{mingdachen,xilun,scottyih\}@meta.com}}

\begin{document}
\maketitle
\begin{abstract}
Few-shot learning for open domain multi-hop question answering typically relies on the in-context learning capability of large language models~(LLMs).
While powerful, these LLMs usually contain tens or hundreds of billions of parameters, making them rather inefficient at inference time.
To improve performance of smaller language models, we propose a data synthesis framework for multi-hop question answering that requires less than 10 human-annotated question answer pairs. Our framework depends only on rich, naturally-occurring relationships among documents and is built upon the data generation functions parameterized by LLMs and prompts. 
We synthesize millions of multi-hop questions and claims to finetune language models, evaluated on popular benchmarks for multi-hop question answering and fact verification.
Empirically, our approach improves model performance significantly, allowing the finetuned models to be competitive with GPT-3.5 based approaches while being almost one-third the size in parameter count.
\end{abstract}
\section{Introduction}
Few-shot learning for open domain multi-hop question answering seeks to answer complex questions by iteratively retrieving relevant information with a handful of human-annotated question answer pairs. It has become increasingly popular for evaluating the abilities of grounding to factual and up-to-date information \citep{lazaridou2022internet} and the reasoning capabilities \citep{press2022measuring} of large language models (LLMs). Recent approaches in this area typically rely on in-context learning \citep{brown2020language} where LLMs are prompted to retrieve relevant information using external search tools \citep{lazaridou2022internet,press2022measuring}. While powerful, the in-context learning capability usually emerges when LLMs have billions of parameters and improves as LLMs become larger in size \citep{wei2022emergent}. This property makes LLMs expensive to experiment with even for inference.

\begin{figure*}
    \centering
    \includegraphics[scale=0.5]{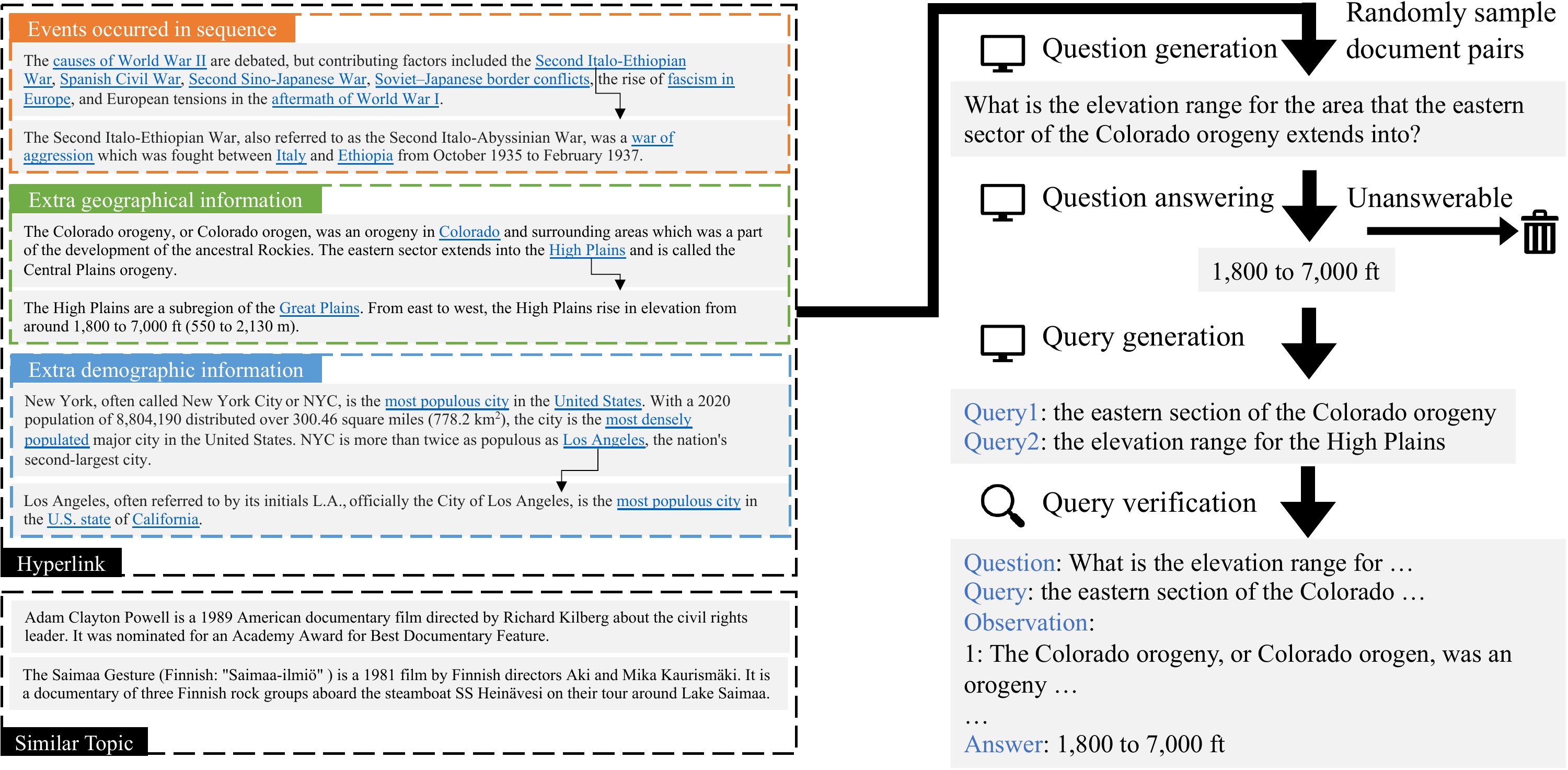}
    \caption{An illustration of the overall pipeline of our proposed approach. Each data instance in our synthesized dataset consists of a question, queries and their corresponding retrieved documents, and an answer. We first prompt LLMs to synthesize questions and queries, finetune models on the synthesized data, and then evaluate the finetuned models on downstream tasks that require iteratively querying retrieval corpora.\vspace{-1em}}
    \label{fig:overall-pipeline}
\end{figure*}

In this work, we propose a data synthesis framework for multi-hop question answering (MQA) that allows for improving smaller language models with less than 10 human-annotated QA pairs (see \Cref{fig:overall-pipeline} for an overall pipeline of our approach). The framework seeks to generate MQA data using documents that are related in different aspects, e.g., sharing similar topics, providing extra information about entities, or talking about events occurred in sequence. This framework is general in that (1) the relationships among documents are naturally-occurring, covering a diverse set of reasoning types; and (2) the data generation pipeline depends on few hand-crafted, task-dependent features.

Specifically, we choose to use Wikipedia as our data sources due to its comprehensive coverage of knowledge and use hyperlinks to capture rich document relationships beyond topic similarity. We start from document pairs that are either topically similar or connected by hyperlinks, then we prompt LLMs to perform three generation tasks: question generation, question answering, and query generation.
We do so by simply changing the format of prompts while re-using the same set of QA pairs. Finally, we verify the quality of queries against retrieval corpora using a neural retriever. We also show that this framework can be easily adapted to other tasks, e.g., fact verification, as demonstrated in our experiments.

Unlike prior work on data synthesis for MQA \citep{pan-etal-2021-unsupervised}, which often depends on carefully designed templates to facilitate complex question generation, limiting the diversity of types of reasoning in their generation questions, our approach requires minimal hand-crafted features as it is built upon LLMs through prompting. In contrast to most work on data synthesis with LLMs \citep[\emph{inter alia}]{schick-schutze-2021-generating,wang-etal-2021-want-reduce} that primarily uses a single data generation function per task, our data generation process involves multiple generation functions because of the complexity of multi-hop question answering.

In experiments, we use a frozen \llamasize{65} \citep{touvron2023llama} to synthesize approximately 1.5 million multi-hop questions and 1.9 million claims, each of which comes with with queries and answers. To validate the effectiveness of the synthetic data, we finetune 7B- and 65B-parameter \llama models on it and then evaluate the finetuned models on three popular multi-hop question answering benchmarks and one fact verification dataset. Empirically, we observe that finetuning on the synthetic data drastically improves model performance, allowing our finetuned \llama 7B to achieve better performance than vanilla \llamasize{65}. Crucially, since the data is synthesized by \llamasize{65}, the improvement from \llamasize{65} essentially comes from the effect similar to self-training.
When comparing to prior work on question and query generation, we show that our approach achieve better performance while requiring less hand-crafted features. Analysis reveals that finetuning on the synthetic data helps models of different sizes, particularly showcasing greater benefits for smaller models. Moreover, we find that automatic filtering steps and having diverse relationships among documents are crucial in improving model performance.

To summarize, our contributions are:

\begin{itemizesquish}
\item We propose a novel data synthesis framework that requires less than 10 human-annotated QA pairs and minimal hand-crafted features;
\item We show that finetuning \llama models on the synthetic data can improve 19.9 points (+63.6\%) and 13.2 points (+33.0\%) on average for the 7B and 65B models respectively. The finetuned \llamasize{7} outperforms the prompting-based \llamasize{65} and finetuned \llamasize{65} achieves results competitive to prior work based on GPT-3.5;
\item We compare to prior work on MQA data generation, demonstrating that our approach achieves better performance while requiring less hand-crafted features.
\end{itemizesquish}

\section{Related Work}

\paragraph{Dataset Synthesis using Language Models.} There have been several attempts in using LLMs to synthesize data for text classification \citep{ye-etal-2022-zerogen,meng2022generating}, semantic similarity predictions \citep{schick-schutze-2021-generating,wang-etal-2021-want-reduce}, question answering \citep{wang-etal-2021-want-reduce,agrawal2022qameleon,ye-etal-2022-zerogen}, summarization \citep{wang-etal-2021-want-reduce}, and instruction tuning \citep{honovich2022unnatural,wang2022self} among others. Unlike these works where they primarily employ one data generation function for a task, our data generation process is built upon a combination of several generation functions due to the complexity of multi-hop question answering. Since our work involves finetuning models on intermediate queries, it is also related to work that finetune models on model-generated intermediate reasoning steps \citep{zelikman2022star,huang2022large,chung2022scaling,yao2023react}. Different from these works, which typically assume the availability of a sizable amount of initial labeled data (e.g., question answer pairs for question answering tasks), our approach requires only a few human annotations.

\paragraph{Question/Query Generation.} Most prior work on automatic multi-hop question generation is cast as a generation task \citep{pan-etal-2020-semantic,su-etal-2020-multi,sachan2020stronger,fei-etal-2022-cqg}, where models are trained in a supervised fashion and designed to maximize the generation metrics, such as BLEU scores \citep{papineni-etal-2002-bleu}. Before prompting LLMs becomes popular, most work attempted to generate queries for information retrieval tasks \citep[\emph{inter alia}]{nogueira2019doc2query,ma-etal-2021-zero,wang-etal-2022-gpl}. In this line of research, \citet{pan-etal-2021-unsupervised} and \citet{qi-etal-2019-answering} are the closest to our work. \citet{pan-etal-2021-unsupervised} try to improve model performance in downstream question answering tasks by augmenting question answer pairs in the training data. \citet{qi-etal-2019-answering} use rule-based algorithms to find overlapping strings between sources and targets to use as queries for multi-hop questions. Although both of these works avoid directly using human supervision, they require heavily hand-crafted data generation functions, and our approach does not. There also are works that automatically generate questions for single-hop question answering \citep{lewis-etal-2021-paq}, language model pretraining \citep{jia-etal-2022-question}, and passage reranking \citep{sachan-etal-2022-improving}.

\paragraph{Prompting for Multi-Hop Question Answering.} \citet{lazaridou2022internet} propose to condition on retrieved information through prompting LLMs.
More recent work prompts LLMs to decompose complex questions into simpler ones through either explicit queries \citep{press2022measuring,yao2023react,khattab2022demonstrate,khot2023decomposed}, integrating retrieval into the chain of thought process \citep{trivedi2022interleaving,jiang2023active}, or sub-questions that can be answered by dedicated question answering models \citep{dua-etal-2022-successive}. \citet{wang-etal-2022-iteratively} and \citet{zhou2023leasttomost} iteratively prompt LLMs to elicit their parametric knowledge. \citet{yoran2023answering} propose to meta-reason over multiple chains of thought instead of using a voting mechanism over the final answers.

\paragraph{Knowledge Distillation.} A large amount of effort has been devoted to distilling smaller models \citep[\emph{inter alia}]{10.1145/1150402.1150464,NIPS2014_ea8fcd92,hinton2015distilling,kim-rush-2016-sequence}. Most recent ones seek to generate datasets \citep{wang-etal-2021-want-reduce} or rationals \citep{wang2023scott,hsieh2023distilling,chen2023zara} from LLMs. However, unlike our work, they either focus on tasks solvable by LLMs' parametric knowledge or assume the availability of amounts of human labeled data. Relatedly, \citet{izacard2021distilling} seek to achieve better performance by distilling knowledge from LLMs to retrievers, whereas in this work, we aim to learn smaller language models and we do not finetune retrievers.
\section{Approach}
We seek to synthesize training data for multi-hop question answering using a handful of human annotations. 
Our data synthesis pipeline leverages naturally-occurring relationships among documents and the powerful reasoning abilities of LLMs.
Each generated data instance contains a question, up to two queries, and an answer. We then finetune models on the generated data.

The data generation process consists of four main steps: question generation, question answering, query generation, and query verification. To achieve this, we use a frozen \llamasize{65} and parameterize the underlying data generation functions with different prompts.\footnote{We will leave the research on further improving model performance by iteratively finetuning on synthetic data and then synthesizing for future work.}

As shown in \Cref{fig:overall-pipeline} Right, our approach can be broken into following steps:

\begin{enumeratesquish}
\item Prepare document pairs
and then randomly choose answers either from context or a pre-defined list of candidates. (\Cref{subsec:data-preparation})\footnote{While our approach can generalize to multiple documents to generate questions with more then two hops, we focus on single- and two-hop questions as prior work found that questions with more than two hops can be difficult to understand even for human readers \citep{press2022measuring}.}
\item Use LLMs to generate questions based on the given documents and answers. (\Cref{subsec:question-generation})
\item Use LLMs to answer the generated questions and only keep those that are answerable. (\Cref{subsec:question-answering})
\item Use LLMs to generate queries given the Wikipedia documents, questions, and answers. (\Cref{subsec:query-generation})
\item Use retrievers to verify the correctness of generated queries against retrieval corpora. (\Cref{subsec:query-verification})
\end{enumeratesquish}

We note that this entire process uses the same set of examples, consisting of up to 10 human-annotated data instances. We use these examples to create prompts for the tasks specified in steps 2, 3, and 4. We describe each step in detail below.

\subsection{Data Preparation}
\label{subsec:data-preparation}
During this step, our objective is to construct data tuples comprising of a pair of documents and an associated answer. To accomplish this, we employ Wikipedia pages as our primary data source, given their comprehensive coverage of knowledge.
We leverage the hyperlinks present within Wikipedia pages, along with the topics of the pages themselves, in order to generate appropriate document pairs.

To extract topics, we finetune a \roberta large model \citep{liu2019roberta} on the DBPedia ontology classification dataset \citep{NIPS2015_250cf8b5} and apply the model to predict the topics of all the Wikipedia pages.\footnote{We use the predicted topics here as opposed to human-annotated category information associated with Wikipedia pages as this approach is more general and can be applied to other data sources without naturally-annotated category information. However, we assume there are abundant data sources for hyperlinks.}
We then cluster documents using the topics. Given a Wikipedia document, we create four document pairs by sampling other documents that either (1) are directly connected by hyperlinks; or (2) belong to the same topic cluster. We will refer to the first setting as ``hyper'' and the second as ``topic''.

We select potential answers in different ways for ``hyper'' and ``topic''. For the ``hyper'' setting, the candidates are from the named entities predicted by the spaCy toolkit and the anchor texts from hyperlinks. For the ``topic'' setting, since generated questions are mostly related to comparing the two documents, we consider the titles of both documents, ``yes'', and ``no'' as candidate answers. We then randomly pick one from the candidate set to use in the final data tuples.

\subsection{Question Generation}
\label{subsec:question-generation}

\begin{figure}[t]
    \centering
    \includegraphics[scale=0.45]{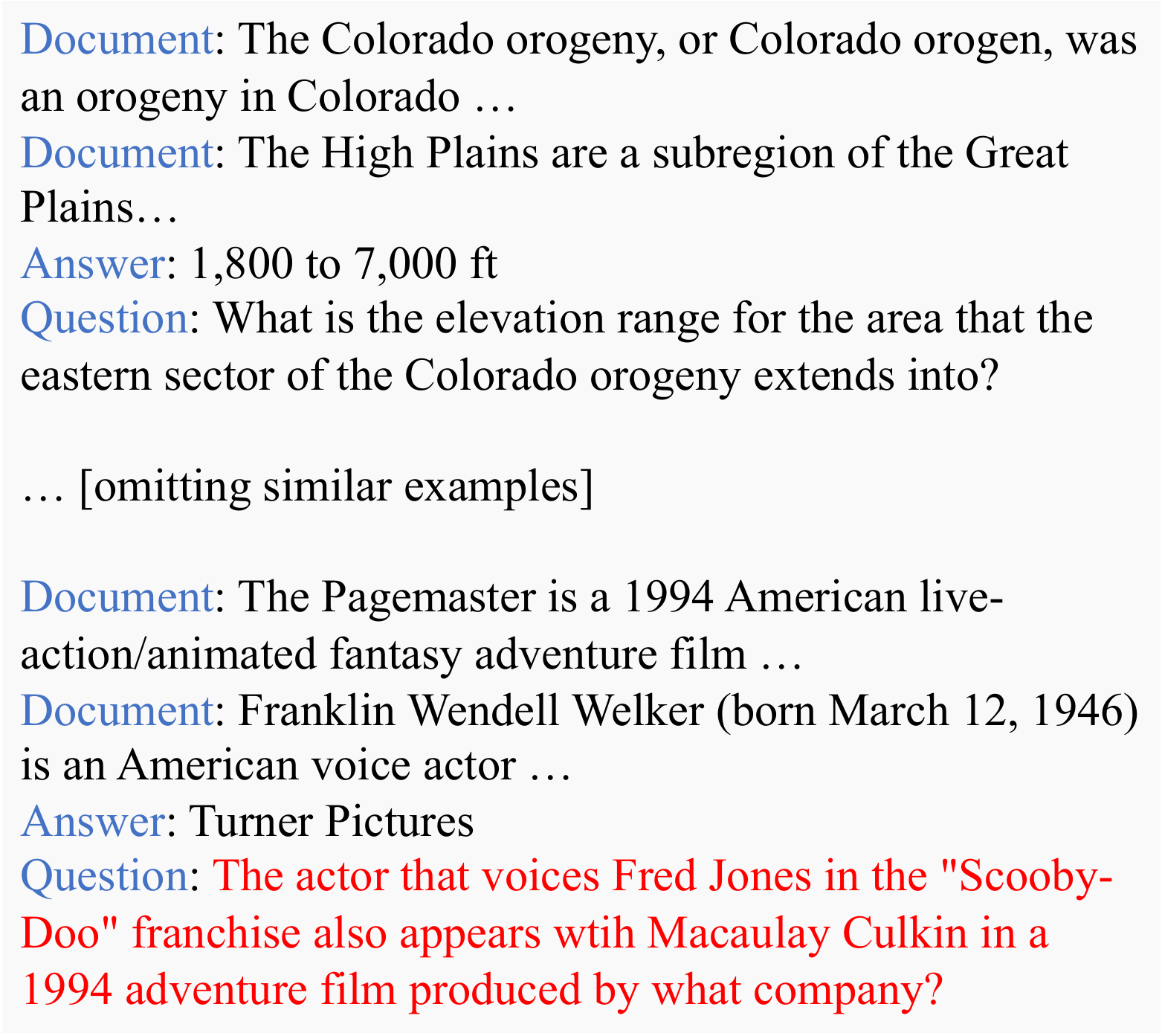}
    \caption{Prompt excerpts for the question generation task for the ``hyper'' setting. The red text is the expected model generation for the given prompt. The complete prompt contains four examples and is included in \Cref{appendix-sec:prompts}.}
    \label{fig:qgen-example}
\end{figure}
As shown in \Cref{fig:qgen-example}, we prompt LLMs to generate questions by providing the prepared document pairs and the associated answer. The examples in the prompt are either from prior work or randomly picked from the training set of \hotpot, consisting of single- and two-hop questions.

Questions generated from the ``topic'' setting are typically related to comparison of two concepts whereas the ones from the ``hyper'' setting tend to be more nested in nature. In light of the different fashions, we use a separate set of examples in the prompts for the ``hyper'' and ``topic'' settings for all of our data generation functions.
We observe LLMs sometimes reference the provided context to ask questions (e.g., What is the birthplace of the man?), which is undesirable since the context will be stripped away when we finetune models on the data. So, we finetune a \roberta large model on the CoNLL-2003 training set \citep{tjong-kim-sang-de-meulder-2003-introduction} to identify named entities in the generated questions. We then drop the questions that have less than one entity in the ``hyper'' setting or less than two entities in the ``topic'' setting. We set the maximum generation step to be 64.

\subsection{Question Answering}
\label{subsec:question-answering}

\begin{figure}[t]
    \centering
    \includegraphics[scale=0.45]{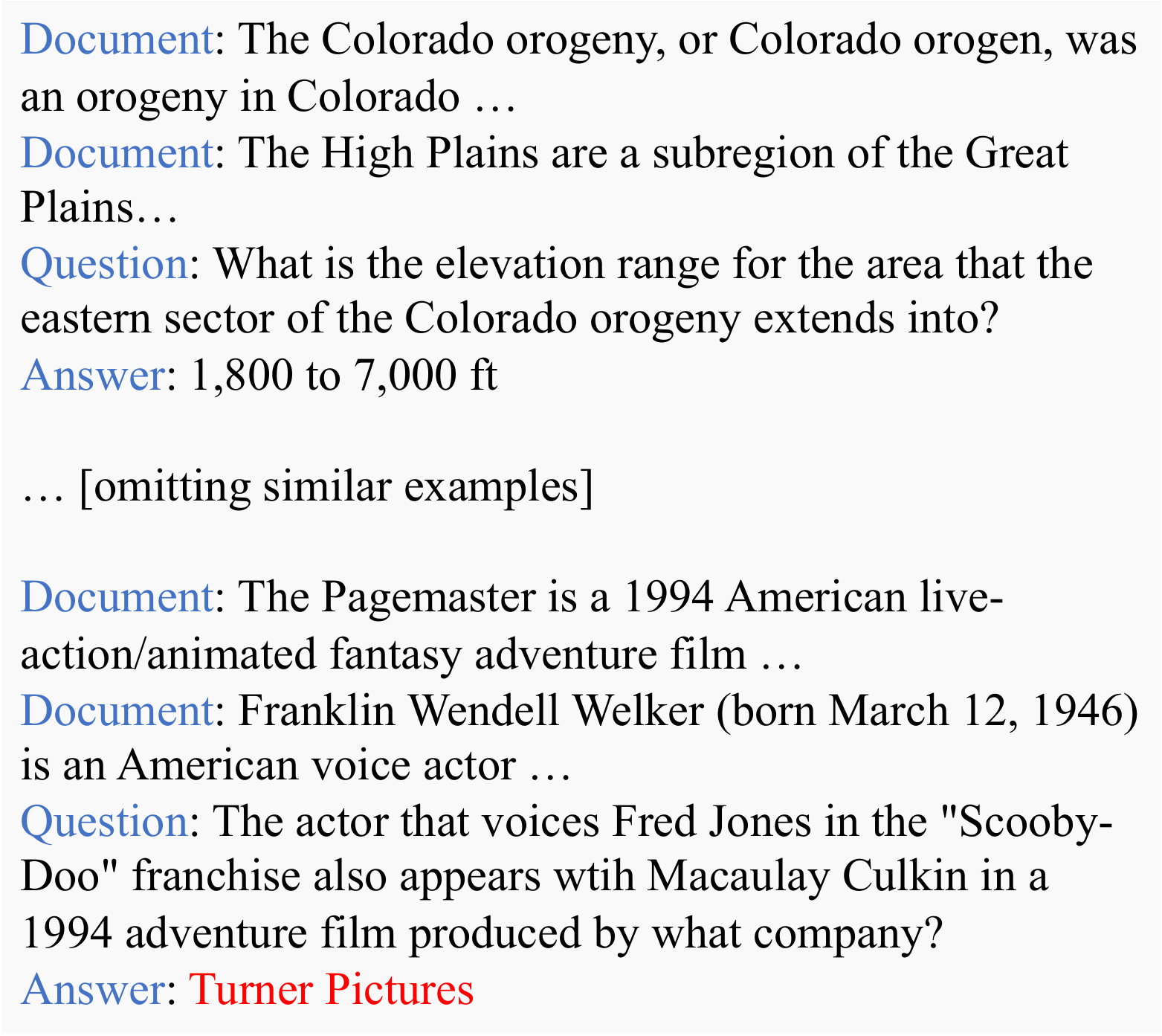}
    \caption{Prompt excerpts for the question answering task for the ``hyper'' setting. The red text is the expected model generation for the given prompt. The complete prompt contains four examples and is included in \Cref{appendix-sec:prompts}.}
    \label{fig:agen-example}
\end{figure}

To verify the correctness of generated questions, we reformat the prompts to ask LLMs to predict answers given the generated questions and the Wikipedia document pairs (see \Cref{fig:agen-example} for an example). We define that a question is ``answerable'' if its LLMs' prediction achieve over 70 $F_1$ scores\footnote{We compute $F_1$ scores by comparing the string of predicted answers to that of ground truth answers after normalization, following \citet{rajpurkar-etal-2016-squad} and \citet{yang-etal-2018-hotpotqa}.} compared to its prepared answer. We set the maximum generation step to be 16.

We also seek to use LLMs to decide whether the questions are single- or two-hop. We do so by prompting LLMs to predict answers when given (1) both documents (``both''); (2) the first document (``first''); and (3) the second document (``second''). We drop questions that are not answerable in ``both''. We keep questions when the prediction from ``both'' agrees with that from either ``second'' or ``first'' even if they differ from the prepared answers. For these questions, we use the predicted answers as ground truths for the rest of experiments. Empirically, we observe this to be a reliable way to increase the amount of synthesized data without sacrificing the quality and these questions are in general single-hop questions.

When deciding the number of hops, we treat all the ``topic'' questions as two-hop questions as they mostly require comparing facts about two concepts, and use the LLMs' predictions to decide the number of hops for ``hyper''. In particular, we classify the ``hyper'' questions that are only answerable in ``both'' as two-hop questions and those that are answerable by ``first'' or ``second'' as single-hop. We will leverage this property later when post-processing generated queries.

\subsection{Query Generation}
\label{subsec:query-generation}
\begin{figure}[t]
    \centering
    \includegraphics[scale=0.45]{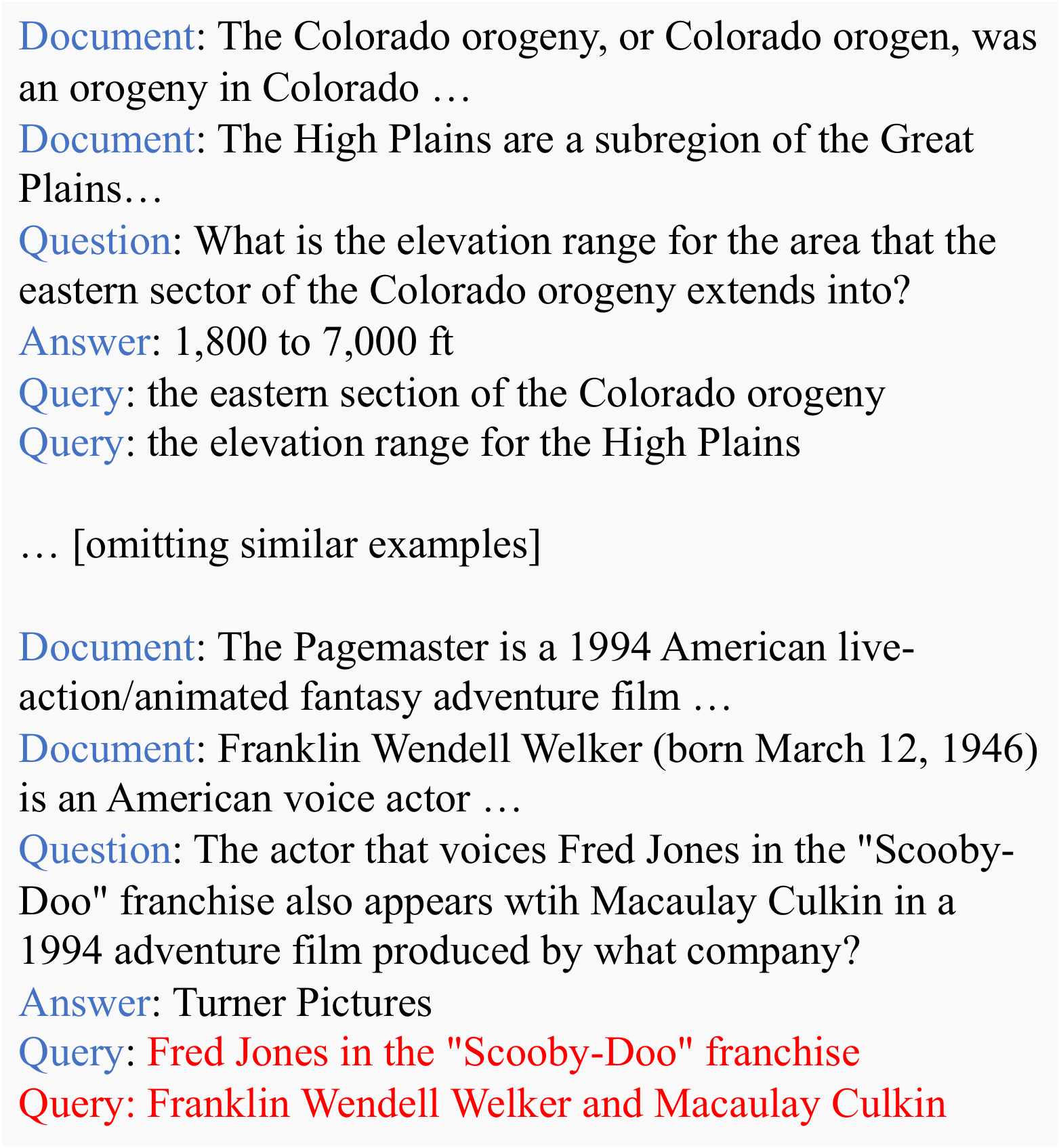}
    \caption{Prompt excerpts for the query generation task for the ``hyper'' setting. The red text is the expected model generation for the given prompt. The complete prompt contains four examples and is included in \Cref{appendix-sec:prompts}.}
    \label{fig:querygen-example}
\end{figure}

As shown in \Cref{fig:querygen-example}, we prompt LLMs to generate retrieval queries given Wikipedia document pairs, generated questions, and the answers from last step. The goal is to generate a sequence of candidate queries, which will later be verified against retrieval corpora using a retriever. We also consider the original question as a candidate query in addition to the model-generated ones. The original questions are used as a backup query at the first hop, i.e., they are included only if the model-generated queries are all classified as invalid in the later verification step. We set the maximum generation step to be 64.

\subsection{Query Verification}
\label{subsec:query-verification}
We take the query candidates and verify whether the queries can retrieve desirable documents from the entire Wikipedia document collections. In this work, we use the \dragon retriever \citep{lin2023train} and the flat index from FAISS \citep{johnson2019billion}.\footnote{Since we only use a subset of Wikipedia documents as retrieval corpus, using flat index is still efficient in our experiments.} We compute similarities among documents using dot product of embedding vectors.

When verifying queries, we seek to find whether a query is valid or a duplicate to another valid query. A query is seen as valid if one of the prepared document pairs is in the top-ranked documents. Queries will be seen as duplicates if they retrieve the same document in the document pair. That is, given a prepared document pair $(d_1, d_2)$, queries $q_1$ and $q_2$, and a retrival function $\text{topk}(\cdot)$ that returns a set of top-ranked documents given a query,
\begin{itemizesquish}
    \item $q_{i}$ is valid if $d_1 \in \text{topk}(q_{i})$ or $d_2 \in \text{topk}(q_{i})$ where $i\in\{1,2\}$;
    \item $q_1$ and $q_2$ are duplicates if $d_1\in \text{topk}(q_1)\cap\text{topk}(q_2)$ or $d_2\in \text{topk}(q_1)\cap\text{topk}(q_2)$.
\end{itemizesquish}

We drop the invalid queries and keep the shortest query if there are duplicates. We also drop questions if we fail to generate valid queries to retrieve (1) both documents for two-hop questions; or (2) the document leading to answerable predictions for single-hop questions (e.g., the first document in the document pair if the questions are answerable in the ``first'' setting). We drop the ``hyper'' questions if their answers are not in the retrieved documents at the last hop. We retrieve top 7 documents in experiments.\footnote{We use 7 documents to ensure enough space to include all these documents without needing to truncate them.}

\subsection{Extend to Fact Verification}
To show that our approach can generalize to other tasks that require multi-hop reasoning, we extend our approach to the fact verification task. We follow the task setup in \fever \citep{thorne-etal-2018-fever} where models are asked to classify whether a claim is ``supported'', ``refuted'', or can not be judged due to ``not enough information''.

In this setting, we also seek to generate a claim, intermediate queries, and an answer. Since facts described in a claim typically come from multiple documents that are closely related, we mostly follow the same procedure as described in previous sections except that we only consider the ``hyper'' document pairs. We use the same prompt for different categories as it improves model performance in our preliminary experiments. We hypothesize that this is due to the fact that \fever is a classification task and providing different task examples within a prompt helps models learn the differences among categories. We use 8 examples in the prompts and show the complete set of prompts in \Cref{appendix-sec:fever-prompts}.
\section{Experiment}
\subsection{Setup}
\begin{table}[t]
\setlength{\tabcolsep}{2pt}
    \centering\small
    \begin{tabular}{l|r|r}
         & Multi-Hop QA & Fact Verification \\\hline
     Size of Train Set& 1,526,266 & 1,985,625 \\
     Size of Dev Set & 5,000 & 5,000 \\
     \#SQ Data & 332,294 (21.7\%) & 1,126,828 (56.7\%) \\
     \#TQ Data & 1,198,972 (78.3\%) & 863,797 (43.3\%) \\\hline
     \multicolumn{3}{l}{\textit{Avgerage number of word tokens}} \\
     Questions/Claims & 14.8 & 10.8 \\
     Queries & 4.4 & 2.6 \\
     Answers & 1.9 & - \\
    \end{tabular}
    \caption{Dataset statistics for synthetic data generated in this work. We omit the average length of answers for fact verification as it is a classification task. SQ=Single-Query. TQ=Two-Queries.}
    \label{tab:synthetic-data-stats}
\end{table}
\paragraph{Training Data.} We synthesize approximately 1.5 million multi-hop questions and 1.9 million claims. We use neucleus sampling \citep{Holtzman2020The} with a top-$p$ probability of 0.9 for decoding when generating the data. Development sets are 5k instances samples from each set. The dataset statistics are summarized in \Cref{tab:synthetic-data-stats}.

\paragraph{Finetuning.} We finetune \llama of two parameter sizes (7B and 65B) on the generated data. During finetuning, we only compute cross-entropy losses on the query and answer strings. We also mix in plain Wikipedia text. Approximately 20\% of data examples in each minibatch are plain text and we finetune \llama on it using vanilla language modeling loss. The finetuning and evaluation experiments are conducted separately for multi-hop QA and fact verification. The best model checkpoints are selected based on the perplexity on the synthesized development sets. We finetune models for 20k steps with a learning rate of 2e-5.

\begin{table}[t]
    \centering\small
    \begin{tabular}{l|c|c|c|c}
         & \hotpot & \musique & \twowiki & \fever \\\hline
    \#data & 7,405 & 1,252 & 12,576 & 19,998 \\
    \#docs& 5,233,328 & 96,720 & 398,354 & 5,396,106 \\
    \end{tabular}
    \caption{Numbers of evaluation data and documents in retrieval corpus used in this work.}
    \label{tab:eval-data-stats}
\end{table}

\begin{table*}[t]
    \centering\small
    \begin{tabular}{l|c|c|c|c|c|c|c|c|c|c}
     & Base & Model  & \multicolumn{2}{c|}{\hotpot} & \multicolumn{2}{c|}{\musique} & \multicolumn{2}{c|}{\twowiki} & \fever & avg. \\ 
     & Model & Size  & EM & $F1$ & EM & $F1$ & EM & $F1$ & Acc & \\\hline
    \multicolumn{8}{l}{\textit{Prior Work}} \\
    ReAct \citep{yao2023react} & PaLM & 540B & 35.1 & - & - & - & - & - & \bf 64.6 & - \\
    SelfAsk \citep{press2022measuring} & GPT-3.5 & 175B & - & - & 15.2 & - & 40.1 & - & - & - \\
    IRCOT \citep{trivedi2022interleaving} & GPT-3.5 & 175B & 50.4 & 61.2 & \bf 31.9 & \bf 42.0 & \bf 53.4 & \bf 65.2 & - & - \\
    DSP \citep{khattab2022demonstrate} & GPT-3.5 & 175B & \bf 51.4 & \bf 62.9 & 24.6 & 36.0 & - & - & - & - \\
    FLARE \citep{jiang2023active} & GPT-3.5 & 175B & - & - & - & - & 51.0 & 59.7 & - & - \\
    MCR \citep{yoran2023answering} & GPT-3.5 & 175B & - & 59.2 & - & - & - & 68.6 & -& - \\\hline
    \multicolumn{8}{l}{\textit{Our Work on \llama 7B}} \\
    SelfAsk$^*$ & \llama & 7B & 16.0 & 22.5 & 4.5 & 11.5 & 24.4 & 28.2 & 34.7 & 22.1 \\
    DSP$^*$ & \llama & 7B & 22.1 & 31.9 & 9.5 & 16.8 & 28.1 & 33.9 & 45.3 & 29.1 \\
    \thiswork & \llama & 7B & 43.0 & 55.2 & 27.2 & 34.7 & 46.3 & 53.2 & 62.9 & 48.2 \\
    \thiswork+ Self-Consistency & \llama & 7B & 44.6 & 56.8 & 28.3 & 35.8 & 46.4 & 53.3 & 63.5 & 49.0 \\\hline
    \multicolumn{8}{l}{\textit{Our Work on \llama 65B}} \\
    SelfAsk$^*$ & \llama & 65B & 35.5 & 46.0 & 20.1 & 28.3 & 35.0 & 42.4 & 50.0 & 30.7 \\
    DSP$^*$ & \llama & 65B & 36.7 & 48.1 & 21.3 & 29.1 & 36.2 & 44.1 & 52.1 & 40.0 \\
    \thiswork & \llama & 65B & 46.4 & 58.6 & 29.6 & 38.6 & 49.3 & 56.6 & 64.1 & 50.9 \\
    \thiswork+ Self-Consistency & \llama & 65B & \bf 49.7 & \bf 62.1 & \bf 31.1 & \bf 41.5 & \bf 51.3 & \bf 60.2 & \bf 65.0 & \bf 53.2 \\
    \end{tabular}
    \caption{Few-shot results on multi-hop question answering and fact verification benchmarks. We list the model size of GPT-3.5 as 175B since prior work uses the DaVinci model, which was estimated to have 175B parameters \citep{openaimodelsize}. We note that the results from prior work are not directly comparable to ours mostly due to the differences in the sizes of evaluation datasets, retrieval corpus, and underlying base models. * indicates our re-implementation. We boldface the best results for GPT-3.5 and our work in each column.}
    \label{tab:main-result}
\end{table*}

\paragraph{Evaluation Benchmarks.} We evaluate finetuned models on three MQA datasets (\hotpot, \musique \citep{trivedi-etal-2022-musique-v2}, and \twowiki \citep{ho-etal-2020-constructing}) and one fact verification datasets (\fever). For all these datasets, we use their entire official development sets as test sets. For \musique, we follow \citet{press2022measuring} to use the subset of two-hop questions. For \fever, we use both the development and test sets in \citet{thorne-etal-2018-fever} as the test set. We report the dataset sizes in \Cref{tab:eval-data-stats}. For multi-hop question answering datasets, we report exact match (EM) and $F_1$ scores. For fact verification, we report accuracies. When averaging scores across datasets, we first take the average of EM and $F_1$ for the MQA datasets and then compute the overall average. Unless otherwise specified we use greedy decoding during evaluation.

\paragraph{Retrieval Corpus.} When generating data, we use the preprocessed Wikipedia dump from \hotpot. For evaluation datasets, we use the preprocessed Wikipedia dumps provided with the datasets for \hotpot and \fever. For \musique and \twowiki, we follow \citet{trivedi2022interleaving} to use all the documents appeared in the datasets as their respective retrieval corpus. We summarize the number of documents for each dataset in \Cref{tab:eval-data-stats}. We note that our retrieval corpus for \musique and \twowiki are smaller than those reported in \citet{trivedi2022interleaving} likely due to the difference in handling duplicate documents, where we simply pick the first document appearing in the datasets. We use the first 100 tokens\footnote{We use spaCy \citep{spacy} tokenizer.} in each Wikipedia page. 

\paragraph{Baselines.} We compare to three kinds of baselines:
\begin{itemizesquish}
\item Prompting based approach: SeflAsk \citep{press2022measuring} and DSP \citep{khattab2022demonstrate}. They are the most competitive few-shot approaches that explicitly issue queries. We re-implement these two approaches using \llama;
\item Prior work on MQA question generation: \citet{pan-etal-2021-unsupervised} heavily rely on hand-crafted functions to ensure the complexity of generated questions;
\item Prior work on query generation for MQA: \citet{qi-etal-2019-answering} use lexical overlap between the retrieval context and the next document(s) expected to retrieve as queries.
\end{itemizesquish}

\subsection{Result}

\paragraph{Compare to prior work on few-shot prompting.} We report our results and the results from prior work in \Cref{tab:main-result}. We apply self-consistency \citep{wang2023selfconsistency}, which samples multiple outputs and then ensembles final predictions based on majority voting, to the finetuned models,\footnote{We use top-$k$ sampling and set the temperature to be 0.7 and $k$ to be 40. We sample 20 outputs per data instance.} which results in additional improvements in model performance. We note that some of prior approaches (e.g., MCR) can be applied to our finetuned models to further improve model performance (e.g., in a way similar to self-consistency).

In general, we find that finetuning on the synthetic data significantly improves model performance for both the 7B- and 65B-parameter \llama. We also observe that \llamasize{7} shows much weaker performance compared to \llamasize{65} when we apply SelfAsk and DSP, which require strong in-context learning capabilities that are often missing in small language models. Interestingly, applying our approach effectively reduces the performance gap between \llamasize{7} and \llamasize{65}. While our results are not directly comparable to those from prior work (due to the differences in evaluation setup), we still include them in the table to show that with our approach \llamasize{65} achieves competitive results than prior work that employs much larger models.

\begin{table}[t]
\setlength{\tabcolsep}{3pt}
    \centering\small
    \begin{tabular}{l|c|c|c|c|c|c|c}
     & \multicolumn{2}{c|}{\hotpot} & \multicolumn{2}{c|}{\musique} & \multicolumn{2}{c}{\twowiki} & avg \\ 
      & EM & $F1$ & EM & $F1$ & EM & $F1$ & \\\hline
      \citet{pan-etal-2021-unsupervised} & 29.9 & 40.3 & 12.2 & 20.4 & 27.0 & 31.8 & 26.9 \\\hline
    \multicolumn{7}{l}{\textit{Our Work}} \\
    Question & 32.7 & 43.4 & 9.9 & 18.4 & 29.4 & 34.5 & 28.1 \\
    Question+Query & \bf 39.2 & \bf 50.7 & \bf 22.3 & \bf 29.8 & \bf 41.1 & \bf 47.8 & \bf 38.5  \\
    \end{tabular}
    \caption{Multi-hop question answering results comparing our work to prior work on few-shot multi-hop question generation. We obtain these results by finetuning \llamasize{7} on 100k data for each setting.}
    \label{tab:question-gen-compare}
\end{table}

\begin{table*}[t]
    \centering\small
    \begin{tabular}{l|c|c|c|c|c|c|c|c|c|c|c|c}
     & \multicolumn{4}{c|}{\hotpot} & \multicolumn{4}{c|}{\musique} & \multicolumn{4}{c}{\twowiki} \\ 
      & EM & $F1$ & prec. & rec. & EM & $F1$ & prec. & rec. & EM & $F1$ & prec. & rec. \\\hline
      \citet{qi-etal-2019-answering} & 31.5 & 42.2 & 55.6 & 55.4 & 15.3 & 23.5 & 55.1 & 46.3 & 32.1 & 35.2 & 71.2 & 66.5 \\
    Our Work &  \bf 39.2 & \bf 50.7 & \bf 81.6 & \bf 69.6 & \bf 22.3 & \bf 29.8 & \bf 64.3 & \bf 57.5 & \bf 41.1 & \bf 47.8 & \bf 93.6 & \bf 80.5 \\
    \end{tabular}
    \caption{Multi-hop question answering results comparing our work to prior work on query generation. We additionally report precision (prec.) and recall (rec.) of the top-ranked documents for each task to measure retrieval performance. We obtain these results by finetuning \llamasize{7} on 100k data for each setting.}
    \label{tab:query-gen-compare}
\end{table*}
\paragraph{Compare to prior work on few-shot multi-hop question generation.} We report results in \Cref{tab:question-gen-compare}. We finetune \llamasize{7} on the 100k questions generated by \citet{pan-etal-2021-unsupervised}.\footnote{Questions are downloaded from the authors' code repository: \url{https://github.com/teacherpeterpan/Unsupervised-Multi-hop-QA}} We also add the few-shot examples that are used to prompt LLMs during our data generation to the training data to ensure fair comparison. As \citet{pan-etal-2021-unsupervised} do not consider intermediate queries, we also finetune \llamasize{7} on 100k questions generated in this work without using queries (``Question''). In both experiments, we retrieve top 15 documents and use the original questions as queries. We find that our generated questions lead to better performance for \hotpot and \twowiki but is worse than \citet{pan-etal-2021-unsupervised} on \musique. Since our approach requires little effort in tuning the data generation functions, these results demonstrate the effectiveness of our approach in generating multi-hop questions. We also experiment with a ``Question+Query'' setting where we finetune models on both questions and their intermediate queries. We observe significant improvements and the final results outperform prior work by a large margin.

\paragraph{Compare to prior work on query generation.}
We adapt the authors' original implementation\footnote{\url{https://github.com/qipeng/golden-retriever}} to generate queries for 100k question answer pairs synthesized by our approach. To measure the retrieval performance, we also report precision and recall for the retrieved documents. In particular, a query prediction is deemed as positive if the ground truth document is within the top-ranked documents. As shown in \Cref{tab:query-gen-compare}, our approach outperforms prior rule-based approach by a significant margin.

\subsection{Analysis}

\begin{figure}[t]
    \centering\small
    \includegraphics[scale=0.3]{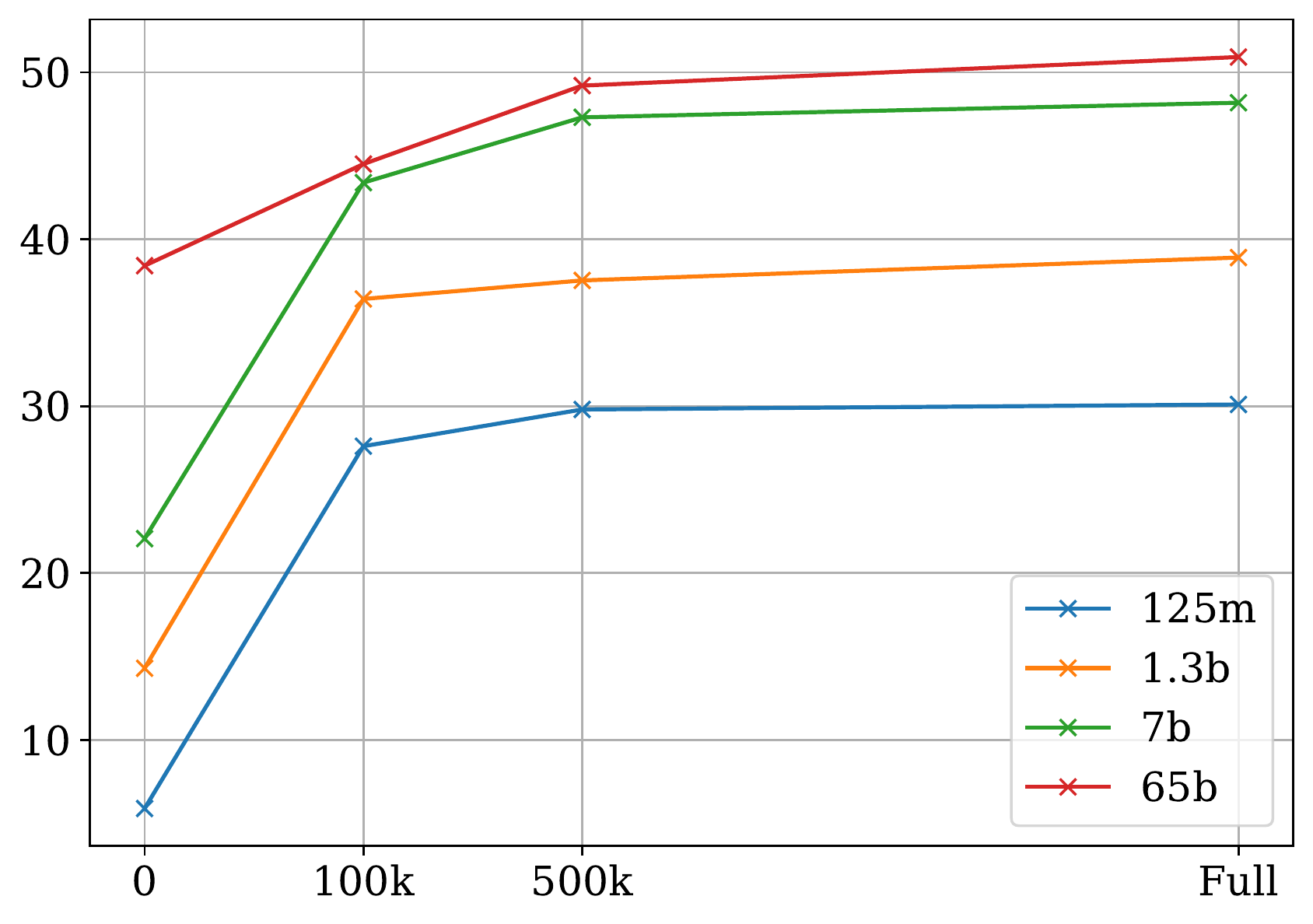}
    \caption{Average dataset performance for \hotpot, \musique, \twowiki, and \fever. We vary the amount of finetuning data and model sizes. We report model performance using SelfAsk when the amount of finetuning data equals to zero.}
    \label{fig:data_model_sizes}
\end{figure}

\paragraph{Effect of Data and Model Sizes.}
To investigate the effect of data and model sizes, we additionally finetune OPT models \citep{zhang2022opt} of 125M and 1.3B parameters on our synthetic datasets, and we vary the amount of the finetuning data (i.e., 0, 100k, 500k, and full). As the general trends are similar across different datasets, we report the average performance for each model when finetuned with a particular amount of data. We note that for multi-hop question answering datasets for which we have two metrics, we take the average of exact match and $F1$ scores as the dataset performance. The results are shown in \Cref{fig:data_model_sizes}. Generally, the synthetic data helps model performance, but larger models still benefit more from the finetuning. The most significant gains are from the initial 100k examples, after which the improvements start to plateau. We will leave the finding of the exact optimal amount of finetuning data for future work.

\begin{table}[t]
\setlength{\tabcolsep}{3pt}
    \centering\small
    \begin{tabular}{l|c|c|c|c|c|c|c}
     & \multicolumn{2}{c|}{\hotpot} & \multicolumn{2}{c|}{\musique} & \multicolumn{2}{c}{\twowiki} & avg \\ 
      & EM & $F1$ & EM & $F1$ & EM & $F1$ &  \\\hline
    QA Pairs & 32.7 & 43.4 & 9.9 & 18.4 & 29.4 & 34.5 & 28.1 \\
    w/o filtering & 21.4 & 22.8 & 4.2 & 10.9 & 22.3 & 26.9 & 18.1  \\
    \hline
    QA Pairs+Queries & 39.2 & 50.7 & 22.3 & 29.8 & 41.1 & 47.8 & 38.5 \\
    w/o filtering & 29.5 & 41.0 & 10.5 & 20.1 & 31.4 & 36.2 & 28.1  \\
    \end{tabular}
    \caption{Results comparing with or without using the filtering steps. We obtain these results by finetuning \llamasize{7} on 100k data for each setting.}
    \label{appendix-tab:filter_step_result}
\end{table}

\paragraph{Effect of Filtering Steps.}
We look into the effect of our filtering steps by finetuning \llamasize{7} models on the unfiltered question answer pairs and unfiltered queries. We report results in \Cref{appendix-tab:filter_step_result}. We note that the filtering step for ``QA Pairs'' corresponds to the question answering step, and the filtering step for the other setting corresponds to the query verification step. In the former setting, similar to previous experiments, we directly retrieve top 15 documents using input questions. In general, we find that the filtering steps help improve model performance significantly.

\paragraph{Effect of Diverse Relationships between Documents.}
\begin{table}[t]
\setlength{\tabcolsep}{3pt}
    \centering\small
    \begin{tabular}{l|c|c|c|c|c|c|c}
     & \multicolumn{2}{c|}{\hotpot} & \multicolumn{2}{c|}{\musique} & \multicolumn{2}{c}{\twowiki} & avg \\ 
      & EM & $F1$ & EM & $F1$ & EM & $F1$ &  \\\hline
    100k hyper + topic & 39.2 & 50.7 & 22.3 & 29.8 & 41.1 & 47.8 & 38.5 \\
    100k hyper & 35.2 & 44.9 & 20.5 & 28.9 & 34.6 & 41.5 & 34.3 \\
    100k topic & 34.9 & 43.8 & 18.9 & 26.8 & 34.8 & 42.1 & 33.6 \\
    \end{tabular}
    \caption{Results when finetuning \llamasize{7} on 100k data which consist of (1) both ``hyper'' and ``topic'' QA pairs, (2) ``hyper'' QA pairs only, and (3) ``topic'' QA pairs only.}
    \label{appendix-tab:diverse_docs_result}
\end{table}

We investigate the effect of finetuning models on data generated from diverse document relationships. We report the results in \Cref{appendix-tab:diverse_docs_result} where we find that diverse document relationships improve multihop QA performance.
\section{Conclusion}
We propose a LLMs-based data synthesis framework for open domain multi-hop question answering that demands less than 10 QA pairs. The framework requires less hand-crafted features than prior work while still achieving better performance. We show that our approach is general by extending to fact verification tasks. Our results on three multi-hop question answering and one fact verification benchmarks show that our approach leads to significantly smaller models that rival the performance of previous methods. The analysis shows (1) the importance of the filtering steps and diverse relationships among documents; and (2) our approach benefits models of various sizes.
\section{Limitations}
We highlight three limitations on our work: (1) our approach depends on synthesizing large amounts of data, which are expensive even if we used \llamasize{65} which are much smaller than PaLM 540B and GPT-3.5; (2) our approach finetunes language models and thus is not applicable to the closed-source language models, e.g., GPT-3 and PaLM; and (3) our approach depends on the availability of powerful LLMs for synthesizing finetuning data.
\bibliography{anthology,custom}

\appendix

\section{Computational Resources}
We use NVIDIA V100’s. It takes approximately 6 GPU hours to generate 1k data points in the final dataset (including the filtering steps). In total, for 3.4 million data points (1.5 million for multi-hop QA and 1.9 million for fact verification) it takes 20.4k GPU hours.

\section{Prompts for Multi-Hop Question Answering}
\label{appendix-sec:prompts}
We show the complete prompts for question generation in \Cref{appendix-tab:qgen-bridge-prompt} and \Cref{appendix-tab:qgen-compare-prompt}. We show the complete prompts for question answering in \Cref{appendix-tab:agen-bridge-prompt} and \Cref{appendix-tab:agen-compare-prompt}. We show the complete prompts for query generation in \Cref{appendix-tab:query-gen-bridge-prompt} and \Cref{appendix-tab:query-gen-compare-prompt}.

\begin{table*}[t]
    \centering\small
    \begin{tabular}{p{0.9\textwidth}}
Document: The Border Surrender were an English rock band based in North London. The band members were Keith Austin (vocals and guitar), Simon Shields (vocals, guitar, bass guitar and mandolin), Johnny Manning (keyboards, melodica, glockenspiel \& accordion) and Mark Austin (drums and vocals).\\
Document: Unsane is an American noise rock trio that was formed in New York City in 1988. Its music touches on elements of hardcore punk and metal.\\
Answer: The Border Surrender\\
Question: Does The Border Surrender or Unsane have more members?\\
\\
Document: Adam Clayton Powell is a 1989 American documentary film directed by Richard Kilberg about the civil rights leader. It was nominated for an Academy Award for Best Documentary Feature.\\
Document: The Saimaa Gesture (Finnish: "Saimaa-ilmiö" ) is a 1981 film by Finnish directors Aki and Mika Kaurismäki. It is a documentary of three Finnish rock groups aboard the steamboat SS Heinävesi on their tour around Lake Saimaa.\\
Answer: The Saimaa Gesture\\
Question: Which documentary is about Finnish rock groups, Adam Clayton Powell or The Saimaa Gesture?\\
\\
Document: Pavel Samuilovich Urysohn (February 3, 1898 - August 17, 1924) was a Soviet mathematician who is best known for his contributions in dimension theory.\\
Document: Leonid Anatolievich Levin is a Soviet-American mathematician and computer scientist.\\ 
Answer: yes\\
Question: Were Pavel Urysohn and Leonid Levin known for the same type of work?\\
\\
Document: Steven Allan Spielberg KBE (born December 18, 1946) is an American film director, writer and producer. He directed Jaws, which is based on the 1974 novel by Peter Benchley.\\
Document: Martin Campbell (born 24 October 1943) is a New Zealand film and television director based in the United Kingdom. He is known for having directed The Mask of Zorro as well as the James Bond films GoldenEye and Casino Royale.\\
Answer: no\\
Question: Are both the directors of Jaws and Casino Royale from the same country? \\

    \end{tabular}
    \caption{Complete prompt for the question generation task in the ``topic'' setting.}
    \label{appendix-tab:qgen-compare-prompt}
\end{table*}

\begin{table*}[t]
    \centering\small
    \begin{tabular}{p{0.9\textwidth}}
Document: The Colorado orogeny, or Colorado orogen, was an orogeny in Colorado and surrounding areas which was a part of the development of the ancestral Rockies. The eastern sector extends into the High Plains and is called the Central Plains orogeny.\\
Document: The High Plains are a subregion of the Great Plains. From east to west, the High Plains rise in elevation from around 1,800 to 7,000 ft (550 to 2,130 m).\\
Answer: 1,800 to 7,000 ft\\
Question: What is the elevation range for the area that the eastern sector of the Colorado orogeny extends into? \\
\\
Document: Avidathe Pole Ivideyum is a 1985 Indian Malayalam drama film directed by K. S. Sethumadhavan and written by John Paul from the story of C. Radhakrishnan. The songs and score were composed by M. K. Arjunan.\\
Document: M. K. Arjunan (1 March 1936 - 6 April 2020) was an Indian film and theatre composer, known for his works in Malayalam cinema and the theatre of Kerala.\\
Answer: 1 March 1936\\
Question: Where was the composer of film Avidathe Pole Ivideyum born?\\
\\
Document: The 1997–98 NBA season was the Pacers' 22nd season in the National Basketball Association. In the off-season, the Pacers hired former Indiana State and Boston Celtics legend Larry Bird as head coach.\\
Document: The 1997–98 NBA season was the 52nd season of the National Basketball Association. The season ended with the Chicago Bulls winning their third straight championship and sixth in the last eight years.\\
Answer: Boston Celtics\\
Question: The head coach during the 1997-98 Indiana Pacers season retired as a player from what NBA team?\\
\\
Document: The Pagemaster is a 1994 American live-action/animated fantasy adventure film starring Macaulay Culkin, Christopher Lloyd, Whoopi Goldberg, Patrick Stewart, Leonard Nimoy, Frank Welker, Ed Begley Jr., and Mel Harris. The film was produced by Turner Pictures.\\
Document: Franklin Wendell Welker (born March 12, 1946) is an American voice actor. Welker is best known for voicing Fred Jones in the Scooby-Doo franchise since its inception in 1969, and the title protagonist himself since 2002.\\
Answer: Turner Pictures\\
Question: The actor that voices Fred Jones in the "Scooby-Doo" franchise also appears wtih Macaulay Culkin in a 1994 adventure film produced by what company?

    \end{tabular}
    \caption{Complete prompt for the question generation task in the ``hyper'' setting.}
    \label{appendix-tab:qgen-bridge-prompt}
\end{table*}

\begin{table*}[t]
    \centering\small
    \begin{tabular}{p{0.9\textwidth}}
Document: The Border Surrender were an English rock band based in North London. The band members were Keith Austin (vocals and guitar), Simon Shields (vocals, guitar, bass guitar and mandolin), Johnny Manning (keyboards, melodica, glockenspiel \& accordion) and Mark Austin (drums and vocals).\\
Document: Unsane is an American noise rock trio that was formed in New York City in 1988. Its music touches on elements of hardcore punk and metal.\\
Question: Does The Border Surrender or Unsane have more members?\\
Answer: The Border Surrender\\
\\
Document: Adam Clayton Powell is a 1989 American documentary film directed by Richard Kilberg about the civil rights leader. It was nominated for an Academy Award for Best Documentary Feature.\\
Document: The Saimaa Gesture (Finnish: "Saimaa-ilmiö" ) is a 1981 film by Finnish directors Aki and Mika Kaurismäki. It is a documentary of three Finnish rock groups aboard the steamboat SS Heinävesi on their tour around Lake Saimaa.\\
Question: Which documentary is about Finnish rock groups, Adam Clayton Powell or The Saimaa Gesture?\\
Answer: The Saimaa Gesture\\
\\
Document: Pavel Samuilovich Urysohn (February 3, 1898 - August 17, 1924) was a Soviet mathematician who is best known for his contributions in dimension theory.\\
Document: Leonid Anatolievich Levin is a Soviet-American mathematician and computer scientist.\\ 
Question: Were Pavel Urysohn and Leonid Levin known for the same type of work?\\
Answer: yes\\
\\
Document: Steven Allan Spielberg KBE (born December 18, 1946) is an American film director, writer and producer. He directed Jaws, which is based on the 1974 novel by Peter Benchley.\\
Document: Martin Campbell (born 24 October 1943) is a New Zealand film and television director based in the United Kingdom. He is known for having directed The Mask of Zorro as well as the James Bond films GoldenEye and Casino Royale.\\
Question: Are both the directors of Jaws and Casino Royale from the same country? \\
Answer: no\\

    \end{tabular}
    \caption{Complete prompt for the question answering task in the ``topic'' setting.}
    \label{appendix-tab:agen-compare-prompt}
\end{table*}

\begin{table*}[t]
    \centering\small
    \begin{tabular}{p{0.9\textwidth}}
Document: The Colorado orogeny, or Colorado orogen, was an orogeny in Colorado and surrounding areas which was a part of the development of the ancestral Rockies. The eastern sector extends into the High Plains and is called the Central Plains orogeny.\\
Document: The High Plains are a subregion of the Great Plains. From east to west, the High Plains rise in elevation from around 1,800 to 7,000 ft (550 to 2,130 m).\\
Question: What is the elevation range for the area that the eastern sector of the Colorado orogeny extends into? \\
Answer: 1,800 to 7,000 ft\\
\\
Document: Avidathe Pole Ivideyum is a 1985 Indian Malayalam drama film directed by K. S. Sethumadhavan and written by John Paul from the story of C. Radhakrishnan. The songs and score were composed by M. K. Arjunan.\\
Document: M. K. Arjunan (1 March 1936 - 6 April 2020) was an Indian film and theatre composer, known for his works in Malayalam cinema and the theatre of Kerala.\\
Question: Where was the composer of film Avidathe Pole Ivideyum born?\\
Answer: 1 March 1936\\
\\
Document: The 1997–98 NBA season was the Pacers' 22nd season in the National Basketball Association. In the off-season, the Pacers hired former Indiana State and Boston Celtics legend Larry Bird as head coach.\\
Document: The 1997–98 NBA season was the 52nd season of the National Basketball Association. The season ended with the Chicago Bulls winning their third straight championship and sixth in the last eight years.\\
Question: The head coach during the 1997-98 Indiana Pacers season retired as a player from what NBA team?\\
Answer: Boston Celtics\\
\\
Document: The Pagemaster is a 1994 American live-action/animated fantasy adventure film starring Macaulay Culkin, Christopher Lloyd, Whoopi Goldberg, Patrick Stewart, Leonard Nimoy, Frank Welker, Ed Begley Jr., and Mel Harris. The film was produced by Turner Pictures.\\
Document: Franklin Wendell Welker (born March 12, 1946) is an American voice actor. Welker is best known for voicing Fred Jones in the Scooby-Doo franchise since its inception in 1969, and the title protagonist himself since 2002.\\
Question: The actor that voices Fred Jones in the "Scooby-Doo" franchise also appears wtih Macaulay Culkin in a 1994 adventure film produced by what company?\\
Answer: Turner Pictures\\

    \end{tabular}
    \caption{Complete prompt for the question answering task in the ``hyper'' setting.}
    \label{appendix-tab:agen-bridge-prompt}
\end{table*}

\begin{table*}[t]
    \centering\small
    \begin{tabular}{p{0.9\textwidth}}
Document: The Border Surrender were an English rock band based in North London. The band members were Keith Austin (vocals and guitar), Simon Shields (vocals, guitar, bass guitar and mandolin), Johnny Manning (keyboards, melodica, glockenspiel \& accordion) and Mark Austin (drums and vocals).\\
Document: Unsane is an American noise rock trio that was formed in New York City in 1988. Its music touches on elements of hardcore punk and metal.\\
Question: Does The Border Surrender or Unsane have more members?\\
Answer: The Border Surrender\\
Query: The Border Surrender \\
Query: Unsane\\
\\
Document: Adam Clayton Powell is a 1989 American documentary film directed by Richard Kilberg about the civil rights leader. It was nominated for an Academy Award for Best Documentary Feature.\\
Document: The Saimaa Gesture (Finnish: "Saimaa-ilmiö" ) is a 1981 film by Finnish directors Aki and Mika Kaurismäki. It is a documentary of three Finnish rock groups aboard the steamboat SS Heinävesi on their tour around Lake Saimaa.\\
Question: Which documentary is about Finnish rock groups, Adam Clayton Powell or The Saimaa Gesture?\\
Answer: The Saimaa Gesture\\
Query: Adam Clayton Powell \\
Query: The Saimaa Gesture \\
\\
Document: Pavel Samuilovich Urysohn (February 3, 1898 - August 17, 1924) was a Soviet mathematician who is best known for his contributions in dimension theory.\\
Document: Leonid Anatolievich Levin is a Soviet-American mathematician and computer scientist.\\ 
Question: Were Pavel Urysohn and Leonid Levin known for the same type of work?\\
Answer: yes\\
Query: Pavel Urysohn \\
Query: Leonid Levin \\
\\
Document: Steven Allan Spielberg KBE (born December 18, 1946) is an American film director, writer and producer. He directed Jaws, which is based on the 1974 novel by Peter Benchley.\\
Document: Martin Campbell (born 24 October 1943) is a New Zealand film and television director based in the United Kingdom. He is known for having directed The Mask of Zorro as well as the James Bond films GoldenEye and Casino Royale.\\
Question: Are both the directors of Jaws and Casino Royale from the same country? \\
Answer: no\\
Query: the director of Jaws \\
Query: the director of Casino Royale \\

    \end{tabular}
    \caption{Complete prompt for the query generation task in the ``topic'' setting.}
    \label{appendix-tab:query-gen-compare-prompt}
\end{table*}

\begin{table*}[t]
    \centering\small
    \begin{tabular}{p{0.9\textwidth}}
Document: The Colorado orogeny, or Colorado orogen, was an orogeny in Colorado and surrounding areas which was a part of the development of the ancestral Rockies. The eastern sector extends into the High Plains and is called the Central Plains orogeny.\\
Document: The High Plains are a subregion of the Great Plains. From east to west, the High Plains rise in elevation from around 1,800 to 7,000 ft (550 to 2,130 m).\\
Question: What is the elevation range for the area that the eastern sector of the Colorado orogeny extends into? \\
Answer: 1,800 to 7,000 ft\\
Query: the eastern section of the Colorado orogeny\\
Query: the elevation range for the High Plains \\
\\
Document: Avidathe Pole Ivideyum is a 1985 Indian Malayalam drama film directed by K. S. Sethumadhavan and written by John Paul from the story of C. Radhakrishnan. The songs and score were composed by M. K. Arjunan.\\
Document: M. K. Arjunan (1 March 1936 - 6 April 2020) was an Indian film and theatre composer, known for his works in Malayalam cinema and the theatre of Kerala.\\
Question: Where was the composer of film Avidathe Pole Ivideyum born?\\
Answer: 1 March 1936\\
Query: the composer of film Avidathe Pole Ivideyum \\
Query: the birthday of M. K. Arjunan\\
\\
Document: The 1997–98 NBA season was the Pacers' 22nd season in the National Basketball Association. In the off-season, the Pacers hired former Indiana State and Boston Celtics legend Larry Bird as head coach.\\
Document: The 1997–98 NBA season was the 52nd season of the National Basketball Association. The season ended with the Chicago Bulls winning their third straight championship and sixth in the last eight years.\\
Question: The head coach during the 1997-98 Indiana Pacers season retired as a player from what NBA team?\\
Answer: Boston Celtics\\
Query: the 1997-98 Indiana Pacers \\
\\
Document: The Pagemaster is a 1994 American live-action/animated fantasy adventure film starring Macaulay Culkin, Christopher Lloyd, Whoopi Goldberg, Patrick Stewart, Leonard Nimoy, Frank Welker, Ed Begley Jr., and Mel Harris. The film was produced by Turner Pictures.\\
Document: Franklin Wendell Welker (born March 12, 1946) is an American voice actor. Welker is best known for voicing Fred Jones in the Scooby-Doo franchise since its inception in 1969, and the title protagonist himself since 2002.\\
Question: The actor that voices Fred Jones in the "Scooby-Doo" franchise also appears wtih Macaulay Culkin in a 1994 adventure film produced by what company?\\
Answer: Turner Pictures\\
Query: Fred Jones in the "Scooby-Doo" franchise \\
Query: Franklin Wendell Welker and Macaulay Culkin \\

    \end{tabular}
    \caption{Complete prompt for the query generation task in the ``hyper'' setting.}
    \label{appendix-tab:query-gen-bridge-prompt}
\end{table*}

\section{Prompts for Fact Verification}
\label{appendix-sec:fever-prompts}

We show the complete prompts used in fact verification in \Cref{appendix-tab:fever-agen-prompt}, \Cref{appendix-tab:fever-qgen-prompt}, and \Cref{appendix-tab:fever-query-gen-prompt}.

\begin{table*}[t]
    \centering\tiny
    \begin{tabular}{p{0.9\textwidth}}
Document: Peggy Sue Got Married is a 1986 American fantasy comedy-drama film directed by Francis Ford Coppola starring Kathleen Turner as a woman on the verge of a divorce, who finds herself transported back to the days of her senior year in high school in 1960.\\
Document: Francis Ford Coppola (born April 7, 1939) is an American film director, producer, and screenwriter. He is considered one of the major figures of the New Hollywood filmmaking movement of the 1960s and 1970s.\\
Claim: Peggy Sue Got Married was one of the most popular films in 1968.\\
Answer: NOT ENOUGH INFO\\
\\
Document: Stranger Things is set in the fictional rural town of Hawkins, Indiana, in the 1980s. The nearby Hawkins National Laboratory ostensibly performs scientific research for the United States Department of Energy but secretly experiments with the paranormal and supernatural, sometimes with human test subjects.\\
Document: Indiana is a U.S. state in the Midwestern United States. It is the 38th-largest by area and the 17th-most populous of the 50 States. Its capital and largest city is Indianapolis.\\
Claim: Stranger Things is set in Bloomington, Indiana.\\
Answer: REFUTES\\
\\
Document: Fort Sumter is a sea fort built on an artificial island protecting Charleston, South Carolina from naval invasion. It was severely damaged during the war, left in ruins, and although there was some rebuilding, the fort as conceived was never completed.\\
Document: Sea forts are completely surrounded by water – if not permanently, then at least at high tide (i.e. they are tidal islands). Unlike most coastal fortifications, which are on the coast, sea forts are not. Instead, they are off the coast on islands, artificial islands, or are specially built structures.\\
Claim: For Sumter was never completed.\\
Answer: SUPPORTS\\
\\
Document: Rodman Edward Serling (December 25, 1924 – June 28, 1975) was an American screenwriter, playwright, television producer, and narrator/on-screen host, best known for his live television dramas of the 1950s and his anthology television series The Twilight Zone. He was known as the "angry young man" of Hollywood, clashing with television executives and sponsors over a wide range of issues, including censorship, racism, and war.\\
Document: The Twilight Zone (marketed as Twilight Zone for its final two seasons) is an American science fiction horror anthology television series created and presented by Rod Serling, which ran for five seasons on CBS from October 2, 1959, to June 19, 1964.\\
Claim: Rod Serling clashed with people.\\
Answer: SUPPORTS\\
\\
Document: Liverpool Football Club is a professional football club based in Liverpool, England. The club competes in the Premier League, the top tier of English football. The club established itself as a major force in domestic and European football in the 1970s and 1980s, when Bill Shankly, Bob Paisley, Joe Fagan and Kenny Dalglish, led the club to a combined 11 League titles and four European Cups.\\
Document: William Shankly OBE (2 September 1913 – 29 September 1981) was a Scottish football player and manager, who is best known for his time as manager of Liverpool. Shankly brought success to Liverpool, gaining promotion to the First Division and winning three League Championships and the UEFA Cup.\\
Claim: Liverpool F.C. did not win a title in 2014.\\
Answer: NOT ENOUGH INFO\\
\\
Document: Nikolaj William Coster-Waldau (born 27 July 1970) is a Danish actor and producer. He played a detective in the short-lived Fox television series New Amsterdam (2008), and appeared in the 2009 Fox television film Virtuality, originally intended as a pilot. \\
Document: The Fox Broadcasting Company, commonly known simply as Fox and stylized in all caps as FOX, is an American commercial broadcast television network owned by Fox Corporation and headquartered in New York City, with master control operations and additional offices at the Fox Network Center in Los Angeles and the Fox Media Center in Tempe.\\
Claim: Nikolaj Coster-Waldau never worked with the Fox Broadcasting Company.\\
Answer: REFUTES\\
\\
Document: X-Men: Days of Future Past is a 2014 American superhero film directed and produced by Bryan Singer and written by Simon Kinberg from a story by Kinberg, Jane Goldman, and Matthew Vaughn. The film is based on the Marvel Comics superhero team The X-Men, the fifth mainline installment of the X-Men film series.\\
Document: The X-Men are a superhero team appearing in American comic books published by Marvel Comics. Created by artist/co-plotter Jack Kirby and writer/editor Stan Lee, the team first appearing in The X-Men \#1 (September 1963).\\
Claim: X-Men: Days of Future Past stars Al Pacino and three cats.\\
Answer: NOT ENOUGH INFO\\
\\
Document: All My Children (often shortened to AMC) is an American television soap opera that aired on ABC from January 5, 1970, to September 23, 2011, and on The Online Network (TOLN) from April 29 to September 2, 2013, via Hulu, Hulu Plus, and iTunes. Created by Agnes Nixon, All My Children is set in Pine Valley, Pennsylvania, a fictional suburb of Philadelphia, which is modeled on the actual Philadelphia suburb of Rosemont.\\
Document: Agnes Nixon (née Eckhardt; December 10, 1922 – September 28, 2016) was an American television writer and producer, and the creator of the ABC soap operas One Life to Live, All My Children, as well as Loving and its spin-off The City.\\
Claim: All My Children was made by a television writer and producer from the United States who passed away in 2016.\\
Answer: SUPPORTS\\

    \end{tabular}
    \caption{Complete prompt for the claim verification task for fact verification.}
    \label{appendix-tab:fever-agen-prompt}
\end{table*}

\begin{table*}[t]
    \centering\tiny
    \begin{tabular}{p{0.9\textwidth}}
Document: Peggy Sue Got Married is a 1986 American fantasy comedy-drama film directed by Francis Ford Coppola starring Kathleen Turner as a woman on the verge of a divorce, who finds herself transported back to the days of her senior year in high school in 1960.\\
Document: Francis Ford Coppola (born April 7, 1939) is an American film director, producer, and screenwriter. He is considered one of the major figures of the New Hollywood filmmaking movement of the 1960s and 1970s.\\
Answer: NOT ENOUGH INFO\\
Claim: Peggy Sue Got Married was one of the most popular films in 1968.\\
\\
Document: Stranger Things is set in the fictional rural town of Hawkins, Indiana, in the 1980s. The nearby Hawkins National Laboratory ostensibly performs scientific research for the United States Department of Energy but secretly experiments with the paranormal and supernatural, sometimes with human test subjects.\\
Document: Indiana is a U.S. state in the Midwestern United States. It is the 38th-largest by area and the 17th-most populous of the 50 States. Its capital and largest city is Indianapolis.\\
Answer: REFUTES\\
Claim: Stranger Things is set in Bloomington, Indiana.\\
\\
Document: Fort Sumter is a sea fort built on an artificial island protecting Charleston, South Carolina from naval invasion. It was severely damaged during the war, left in ruins, and although there was some rebuilding, the fort as conceived was never completed.\\
Document: Sea forts are completely surrounded by water – if not permanently, then at least at high tide (i.e. they are tidal islands). Unlike most coastal fortifications, which are on the coast, sea forts are not. Instead, they are off the coast on islands, artificial islands, or are specially built structures.\\
Answer: SUPPORTS\\
Claim: For Sumter was never completed.\\
\\
Document: Rodman Edward Serling (December 25, 1924 – June 28, 1975) was an American screenwriter, playwright, television producer, and narrator/on-screen host, best known for his live television dramas of the 1950s and his anthology television series The Twilight Zone. He was known as the "angry young man" of Hollywood, clashing with television executives and sponsors over a wide range of issues, including censorship, racism, and war.\\
Document: The Twilight Zone (marketed as Twilight Zone for its final two seasons) is an American science fiction horror anthology television series created and presented by Rod Serling, which ran for five seasons on CBS from October 2, 1959, to June 19, 1964.\\
Answer: SUPPORTS\\
Claim: Rod Serling clashed with people.\\
\\
Document: Liverpool Football Club is a professional football club based in Liverpool, England. The club competes in the Premier League, the top tier of English football. The club established itself as a major force in domestic and European football in the 1970s and 1980s, when Bill Shankly, Bob Paisley, Joe Fagan and Kenny Dalglish, led the club to a combined 11 League titles and four European Cups.\\
Document: William Shankly OBE (2 September 1913 – 29 September 1981) was a Scottish football player and manager, who is best known for his time as manager of Liverpool. Shankly brought success to Liverpool, gaining promotion to the First Division and winning three League Championships and the UEFA Cup.\\
Answer: NOT ENOUGH INFO\\
Claim: Liverpool F.C. did not win a title in 2014.\\
\\
Document: Nikolaj William Coster-Waldau (born 27 July 1970) is a Danish actor and producer. He played a detective in the short-lived Fox television series New Amsterdam (2008), and appeared in the 2009 Fox television film Virtuality, originally intended as a pilot. \\
Document: The Fox Broadcasting Company, commonly known simply as Fox and stylized in all caps as FOX, is an American commercial broadcast television network owned by Fox Corporation and headquartered in New York City, with master control operations and additional offices at the Fox Network Center in Los Angeles and the Fox Media Center in Tempe.\\
Answer: REFUTES\\
Claim: Nikolaj Coster-Waldau never worked with the Fox Broadcasting Company.\\
\\
Document: X-Men: Days of Future Past is a 2014 American superhero film directed and produced by Bryan Singer and written by Simon Kinberg from a story by Kinberg, Jane Goldman, and Matthew Vaughn. The film is based on the Marvel Comics superhero team The X-Men, the fifth mainline installment of the X-Men film series.\\
Document: The X-Men are a superhero team appearing in American comic books published by Marvel Comics. Created by artist/co-plotter Jack Kirby and writer/editor Stan Lee, the team first appearing in The X-Men \#1 (September 1963).\\
Answer: NOT ENOUGH INFO\\
Claim: X-Men: Days of Future Past stars Al Pacino and three cats.\\
\\
Document: All My Children (often shortened to AMC) is an American television soap opera that aired on ABC from January 5, 1970, to September 23, 2011, and on The Online Network (TOLN) from April 29 to September 2, 2013, via Hulu, Hulu Plus, and iTunes. Created by Agnes Nixon, All My Children is set in Pine Valley, Pennsylvania, a fictional suburb of Philadelphia, which is modeled on the actual Philadelphia suburb of Rosemont.\\
Document: Agnes Nixon (née Eckhardt; December 10, 1922 – September 28, 2016) was an American television writer and producer, and the creator of the ABC soap operas One Life to Live, All My Children, as well as Loving and its spin-off The City.\\
Answer: SUPPORTS\\
Claim: All My Children was made by a television writer and producer from the United States who passed away in 2016.\\

    \end{tabular}
    \caption{Complete prompt for the claim generation task for fact verification.}
    \label{appendix-tab:fever-qgen-prompt}
\end{table*}

\begin{table*}[t]
    \centering\tiny
    \begin{tabular}{p{0.9\textwidth}}
Document: Peggy Sue Got Married is a 1986 American fantasy comedy-drama film directed by Francis Ford Coppola starring Kathleen Turner as a woman on the verge of a divorce, who finds herself transported back to the days of her senior year in high school in 1960.\\
Document: Francis Ford Coppola (born April 7, 1939) is an American film director, producer, and screenwriter. He is considered one of the major figures of the New Hollywood filmmaking movement of the 1960s and 1970s.\\
Claim: Peggy Sue Got Married was one of the most popular films in 1968.\\
Answer: NOT ENOUGH INFO\\
Query: Peggy Sue Got Married\\
\\
Document: Stranger Things is set in the fictional rural town of Hawkins, Indiana, in the 1980s. The nearby Hawkins National Laboratory ostensibly performs scientific research for the United States Department of Energy but secretly experiments with the paranormal and supernatural, sometimes with human test subjects.\\
Document: Indiana is a U.S. state in the Midwestern United States. It is the 38th-largest by area and the 17th-most populous of the 50 States. Its capital and largest city is Indianapolis.\\
Claim: Stranger Things is set in Bloomington, Indiana.\\
Answer: REFUTES\\
Query: Stranger Things\\
\\
Document: Fort Sumter is a sea fort built on an artificial island protecting Charleston, South Carolina from naval invasion. It was severely damaged during the war, left in ruins, and although there was some rebuilding, the fort as conceived was never completed.\\
Document: Sea forts are completely surrounded by water – if not permanently, then at least at high tide (i.e. they are tidal islands). Unlike most coastal fortifications, which are on the coast, sea forts are not. Instead, they are off the coast on islands, artificial islands, or are specially built structures.\\
Claim: For Sumter was never completed.\\
Answer: SUPPORTS\\
Query: For Sumter\\
\\
Document: Rodman Edward Serling (December 25, 1924 – June 28, 1975) was an American screenwriter, playwright, television producer, and narrator/on-screen host, best known for his live television dramas of the 1950s and his anthology television series The Twilight Zone. He was known as the "angry young man" of Hollywood, clashing with television executives and sponsors over a wide range of issues, including censorship, racism, and war.\\
Document: The Twilight Zone (marketed as Twilight Zone for its final two seasons) is an American science fiction horror anthology television series created and presented by Rod Serling, which ran for five seasons on CBS from October 2, 1959, to June 19, 1964.
Claim: Rod Serling clashed with people.\\
Answer: SUPPORTS\\
Query: Rod Serling\\
\\
Document: Liverpool Football Club is a professional football club based in Liverpool, England. The club competes in the Premier League, the top tier of English football. The club established itself as a major force in domestic and European football in the 1970s and 1980s, when Bill Shankly, Bob Paisley, Joe Fagan and Kenny Dalglish, led the club to a combined 11 League titles and four European Cups.\\
Document: William Shankly OBE (2 September 1913 – 29 September 1981) was a Scottish football player and manager, who is best known for his time as manager of Liverpool. Shankly brought success to Liverpool, gaining promotion to the First Division and winning three League Championships and the UEFA Cup.\\
Claim: Liverpool F.C. did not win a title in 2014.\\
Answer: NOT ENOUGH INFO\\
Query: Liverpool F.C.\\
\\
Document: Nikolaj William Coster-Waldau (born 27 July 1970) is a Danish actor and producer. He played a detective in the short-lived Fox television series New Amsterdam (2008), and appeared in the 2009 Fox television film Virtuality, originally intended as a pilot. \\
Document: The Fox Broadcasting Company, commonly known simply as Fox and stylized in all caps as FOX, is an American commercial broadcast television network owned by Fox Corporation and headquartered in New York City, with master control operations and additional offices at the Fox Network Center in Los Angeles and the Fox Media Center in Tempe.\\
Claim: Nikolaj Coster-Waldau never worked with the Fox Broadcasting Company.\\
Answer: REFUTES\\
Query: Nikolaj Coster-Waldau\\
Query: Fox television\\
\\
Document: X-Men: Days of Future Past is a 2014 American superhero film directed and produced by Bryan Singer and written by Simon Kinberg from a story by Kinberg, Jane Goldman, and Matthew Vaughn. The film is based on the Marvel Comics superhero team The X-Men, the fifth mainline installment of the X-Men film series.\\
Document: The X-Men are a superhero team appearing in American comic books published by Marvel Comics. Created by artist/co-plotter Jack Kirby and writer/editor Stan Lee, the team first appearing in The X-Men \#1 (September 1963).\\
Claim: X-Men: Days of Future Past stars Al Pacino and three cats.\\
Answer: NOT ENOUGH INFO\\
Query: X-Men: Days of Future Past\\
\\
Document: All My Children (often shortened to AMC) is an American television soap opera that aired on ABC from January 5, 1970, to September 23, 2011, and on The Online Network (TOLN) from April 29 to September 2, 2013, via Hulu, Hulu Plus, and iTunes. Created by Agnes Nixon, All My Children is set in Pine Valley, Pennsylvania, a fictional suburb of Philadelphia, which is modeled on the actual Philadelphia suburb of Rosemont.\\
Document: Agnes Nixon (née Eckhardt; December 10, 1922 – September 28, 2016) was an American television writer and producer, and the creator of the ABC soap operas One Life to Live, All My Children, as well as Loving and its spin-off The City.\\
Claim: All My Children was made by a television writer and producer from the United States who passed away in 2016.\\
Answer: SUPPORTS\\
Query: All My Children\\
Query: Agnes Nixon\\

    \end{tabular}
    \caption{Complete prompt for the query generation task for fact verification.}
    \label{appendix-tab:fever-query-gen-prompt}
\end{table*}

\end{document}